\documentclass[sigconf]{acmart}
\usepackage{algorithm}
\usepackage{algpseudocode}
\usepackage{graphicx} % For including graphics and resizing boxes
\usepackage{multirow} % For multi-row cells in tables
\usepackage{booktabs} % For better table formatting

\renewcommand\footnotetextcopyrightpermission[1]{}
\titlenote{Copyright 2024 for this paper by its authors. Use permitted under Creative Commons License Attribution 4.0 International (CC BY 4.0).\\Presented at the SURE workshop held in conjunction with the 18th ACM Conference on Recommender Systems (RecSys), 2024, in Bari, Italy.}

\acmConference[Conference'24]{ACM Conference}{October 2024}{Bari, Italy}

\AtBeginDocument{%
  }

\setcopyright{acmlicensed}
\copyrightyear{2024}
\acmYear{2024}
\acmDOI{XXXXXXX.XXXXXXX}

\acmJournal{TOG}
\acmVolume{37}
\acmNumber{4}
\acmArticle{111}
\acmMonth{8}

\begin{document}

\title{Beauty Beyond Words: Explainable Beauty Product Recommendations Using Ingredient-Based Product Attributes}


\author{Celine Liu}
\email{celineli@amazon.com}
\affiliation{%
  \institution{Amazon}
  \city{Vancouver}
  % \state{BC}
  \country{Canada}
}

\author{Rahul Suresh}
\email{surerahu@amazon.com}
\affiliation{%
  \institution{Amazon}
  \city{Vancouver}
  % \state{BC}
  \country{Canada}
}

\author{Amin Banitalebi-Dehkordi}
\email{aminbt@amazon.com}
\affiliation{%
  \institution{Amazon}
  \city{Vancouver}
  % \state{BC}
  \country{Canada}
}

\begin{abstract}

Accurate attribute extraction is critical for beauty product recommendations and building trust with customers. This remains an open problem, as existing solutions are often unreliable and incomplete. We present a system to extract beauty-specific attributes using end-to-end supervised learning based on beauty product ingredients. A key insight to our system is a novel energy-based implicit model architecture. We show that this implicit model architecture offers significant benefits in terms of accuracy, explainability, robustness, and flexibility. Furthermore, our implicit model can be easily fine-tuned to incorporate additional attributes as they become available, making it more useful in real-world applications. We validate our model on a major e-commerce skincare product catalog dataset and demonstrate its effectiveness. Finally, we showcase how ingredient-based attribute extraction contributes to enhancing the explainability of beauty recommendations.

\end{abstract}

%
% Keywords. The author(s) should pick words that accurately describe
% the work being presented. Separate the keywords with commas.


%   \includegraphics[width=\textwidth]{Figures/widget.png}
%   \caption{figure caption}
%   % \Description{figure description}
\maketitle

% \section{Introduction}
% ACM's consolidated article template, introduced in 2017, provides a
% consistent \LaTeX\ style for use across ACM publications, and
% incorporates accessibility and metadata-extraction functionality
% necessary for future Digital Library endeavors. Numerous ACM and
% SIG-specific \LaTeX\ templates have been examined, and their unique
% features incorporated into this single new template.

% If you are new to publishing with ACM, this document is a valuable
% guide to the process of preparing your work for publication. If you
% have published with ACM before, this document provides insight and
% instruction into more recent changes to the article template.

% The ``\verb|acmart|'' document class can be used to prepare articles
% for any ACM publication --- conference or journal, and for any stage
% of publication, from review to final ``camera-ready'' copy, to the
% author's own version, with {\itshape very} few changes to the source.

% \input{Sections/problem}
\section{Introduction}

\begin{figure*}
    \centering
    \includegraphics[width=0.8\linewidth]{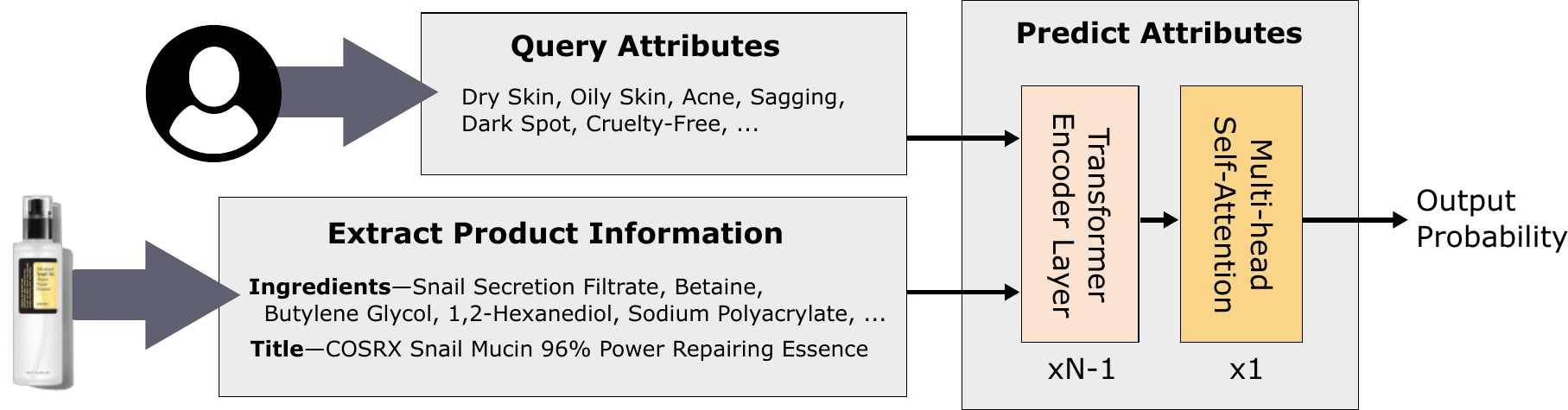}
    \caption{Overview of beauty product extraction workflow and the BT-BERT architecture. Our model is identical to the BERT Transformer~\cite{bert} except in the last layer---the initial \texttt{N-1} layers remain unmodified. We remove the final MLP from the last layer of the Transformer encoder and directly use the self-attention values to formulate the output probability.}
    \label{fig:implicit_diagram}
\end{figure*}

%introduction should also have some details on what is wrong with exlicit models
The value of the global beauty and personal care market is estimated to be over \$646 billion in 2024~\cite{business-value-1}.
Product discovery and trust are two of the biggest considerations in Beauty customers’ shopping journeys in e-commerce stores.
Many factors contribute to these problems, such as lack of personalized recommendations, inaccurate or incomplete product benefit and/or ingredient information, lack of targeted curation, etc. 
Having such information accurately listed in the product catalogue is particularly important for Beauty category of products, as they are topically applied to the skin. 
Manual curation and sanitization of such metadata is possible at small scales. 
However, for larger e-commerce stores, with a large portfolio of products, it will be impractical to rely on manual annotation.

The primary objective of our work is to enhance the beauty shopping experience by automatically and accurately extracting beauty attributes at scale. 
These attributes not only aid customers in comparing and refining product choices but also foster trust in the e-commerce stores. 
Furthermore, the extracted attributes contribute to building more explainable beauty recommendations, which empower customers to make informed purchasing decisions.

We propose a robust and scalable learning-based solution capable of predicting beauty attributes from product ingredients.
To achieve this, we integrate an energy-based implicit strategy to extract 5 skin types, 11 skin concerns, and 17 attributes commonly preferred across beauty products, as elaborated in \autoref{sec:label_category}. 
In summary, the key benefits of our proposed model are:
\begin{itemize}
    
    \item Improved \textbf{accuracy} and \textbf{precision} compared to the alternatives,
    \item \textbf{Explainability} through analysis of the attention weights (\hyperref[sec:explainability]{\S\ref*{sec:explainability}}),
    \item \textbf{Robustness} in a low-resource regime via implicit data augmentation (\hyperref[sec:robustness]{\S\ref*{sec:robustness}}),
    \item \textbf{Flexibility} when finetuning previously trained models on new labels (\hyperref[sec:fine-tune]{\S\ref*{sec:fine-tune}}).
\end{itemize}

To the best of our knowledge, there has been no prior study on the extraction of beauty-specific attributes based on product ingredients. Our contributions are outlined as follows:
\begin{itemize}
\item We introduce a novel energy-based implicit model for extracting beauty attributes from product ingredients and the title. We define \textit{implicit} vs.~\textit{explicit} models in \autoref{sec:system_overview}.
% \item Our proposed approach is assessed using skincare products from Amazon.com. We demonstrate its superiority over traditional keyword-based solutions and an explicit classifier baseline on a test dataset annotated by beauty domain experts.
\item Our proposed approach is assessed using skincare products from a major e-commerce store. We demonstrate its superiority over traditional keyword-based solutions and an explicit classifier baseline on a test dataset annotated by beauty domain experts.
% \item We document and extensively discuss the key algorithmic and architectural features that contribute to explainability, robustness, and flexibility of our proposed model. We verify the correctness of the extracted attributes by constructing an implicit customer profile from purchase history and comparing the results to a user survey~\autoref{sec:customer-segmentation}.
\item We document and extensively discuss the key algorithmic and architectural features that contribute to explainability, robustness, and flexibility of our proposed model.
% \item We make our meticulously curated dataset available to facilitate future research\footnote{Available upon request}.
% \item \note{As a use-case study, we show how the extracted attributes can be used for customer implicit profile creation, that can help with customer segmentation and a more personalized recommendations, in future.}
% \item As a use-case study, we show how the extracted attributes can be used for customer understanding and a more personalized recommendations in ~\autoref{sec:customer-segmentation}.
\item As a use-case study, we illustrate how ingredient-based extracted attributes can enhance the development of explainable beauty recommendations in ~\autoref{sec:customer-segmentation}.
\end{itemize}

\section{Related Works}
\label{sec:related-work}

%%%%%%%%%%%%% regarding attribute extraction task
% NER
\paragraph{\bf Attribute Value Extraction} The problem of product attribute extraction in e-commerce is traditionally solved using named entity recognition (NER).
NER approaches typically use beginning-inside-outside (BIO) tagging~\cite{domain-2010, bootstrapped-NER-2011} to segment texts. 
% However, NER-based approaches have limitations as they depend on predefined entity types, and thus making them challenging to scale in environments with thousands of attributes and constant changes. 
However, NER-based approaches exhibit substantial limitations due to their reliance on predefined entity types. 
This rigidity makes it difficult to scale in dynamic environments where attributes are numerous and constantly evolving, such as in beauty product recommendations. 
% This approach also lacks the flexibility to dynamically model the relationships between attributes and their values, necessitating substantial manual effort to accommodate new data.
% Moreover, NER struggles to handle ambiguous terms or adapt to new attributes effectively. 
% Its inflexible structure also fails to dynamically model the intricate relationships between attributes and their values, often necessitating extensive manual intervention to integrate new data effectively.
% deep learning + sequence problem
Certain research also models the attribute extraction task as a sequential tagging problem ~\cite{bilstm-crf, opentag} using CRF and BiLSTM.
% ~\cite{adatag} extracts the attributes with a parameterized decoder with pretrained attribute embeddings through a hypernetwork and a Mixture-of-Experts (MoE) module. 
~\cite{adatag} describes a method that extracts attributes using a parameterized decoder with pretrained attribute embeddings, through a hypernetwork and a Mixture-of-Experts (MoE) module.
~\cite{xu2019scaling} also model the attribute to make the prediction task more scalable.  
Our work is similar to the solution proposed in~\cite{xu2019scaling}, which uses BERT and Bi-LSTMs to model semantic relations between attribute and product titles on a large-scale dataset.
However, the deep learning modules in~\cite{xu2019scaling} are primarily used as components in the NER pipeline and the outputs of the model are still the BIO tags.
Our work is different in that our proposed model directly outputs the attribute values and the architectural design choices are heavily guided by explainability, robustness, and flexibility.

% multimodality and multitask
% More recent advancement for product attribute extraction 
In the direction of classification tasks, recent advancements utilize multitask framework and multi-modality~\cite{cardoso2018product, wang-etal-2022-smartave, dezaki2023automated}.
Furthermore, these models utilize parameter sharing across different attribute prediction tasks, reducing the model's complexity and encouraging generalization.
Each attribute has its own output layer, allowing the network to predict multiple attributes simultaneously.
%%%%%%%%%%%%% regarding implicit classification task
On the other hand, prior works have demonstrated that incorporating an implicit method~\cite{ebm-2019, implicit-2021} offers unique benefits. 
In particular, when treating product attribute extraction as an implicit classification problem---where attributes themselves are also part of the input---the model can focus on specific attributes to extract from the product description. 
This approach helps the model learn more meaningful and relevant embeddings from the input which leads to more accurate attribute value extraction.
% Our work is similar to the solution proposed in~\cite{xu2019scaling}, which uses BERT and Bi-LSTMs to model semantic relations between attribute and product titles on a large scale dataset.
% However, the deep learning modules in~\cite{xu2019scaling} are primarily used as components in the NER pipeline and the outputs of the model are still the BIO tags.
% Our work is different in that our proposed model directly outputs the attribute values and the architectural design choices are heavily guided by explainability, robustness, and flexibility.

% It's also important to mention that previous work has approached attribute extraction as a means of backfilling missing data. 
% This involves using existing data to train a model to predict additional missing attributes. 
% However, in the beauty industry, ensuring attribute accuracy is crucial. 
% For this reason, our training data is meticulously and scientifically labeled by experts in the field.

%%%%%%%%%%%%% regarding personalization 
\paragraph{\bf Beauty Product Recommendation} Extant literature provides limited research on beauty product recommendation that incorporates ingredient analysis~\cite{afshar2023improving, alashkar2017examples}. 
~\cite{9394031} directly uses an ingredient-concern mapping table to provide solutions for users of various skin conditions detected by an object detection computer vision model. 
However, this mapping table is often supplied by a third party where mappings are constructed independently for each ingredient without accounting for the order and the interactions with other ingredients, leading to inflexible rule-based recommendation methods.
~\cite{8912051}'s approach extracts ingredient efficacy based on user reviews and recommends products containing those ingredients for customers across various age groups. 
Although this method relies on user-generated content, it does not align with our fact-based approach, making it inapplicable to our use-case scenario.
~\cite{9767972} employs a method based on ingredient similarity using one-hot encoding to recommend products given a user's past purchase. 
However, this work does not leverage ingredient data to predict targeted skin types and concerns directly, which is the focus of our work.

\section{System Overview}
\label{sec:system_overview}
%I think more details are needed. We should have convincing explanation of why this implicit model is better? 

\begin{figure*}
    \centering
    \includegraphics[width=0.85\textwidth]{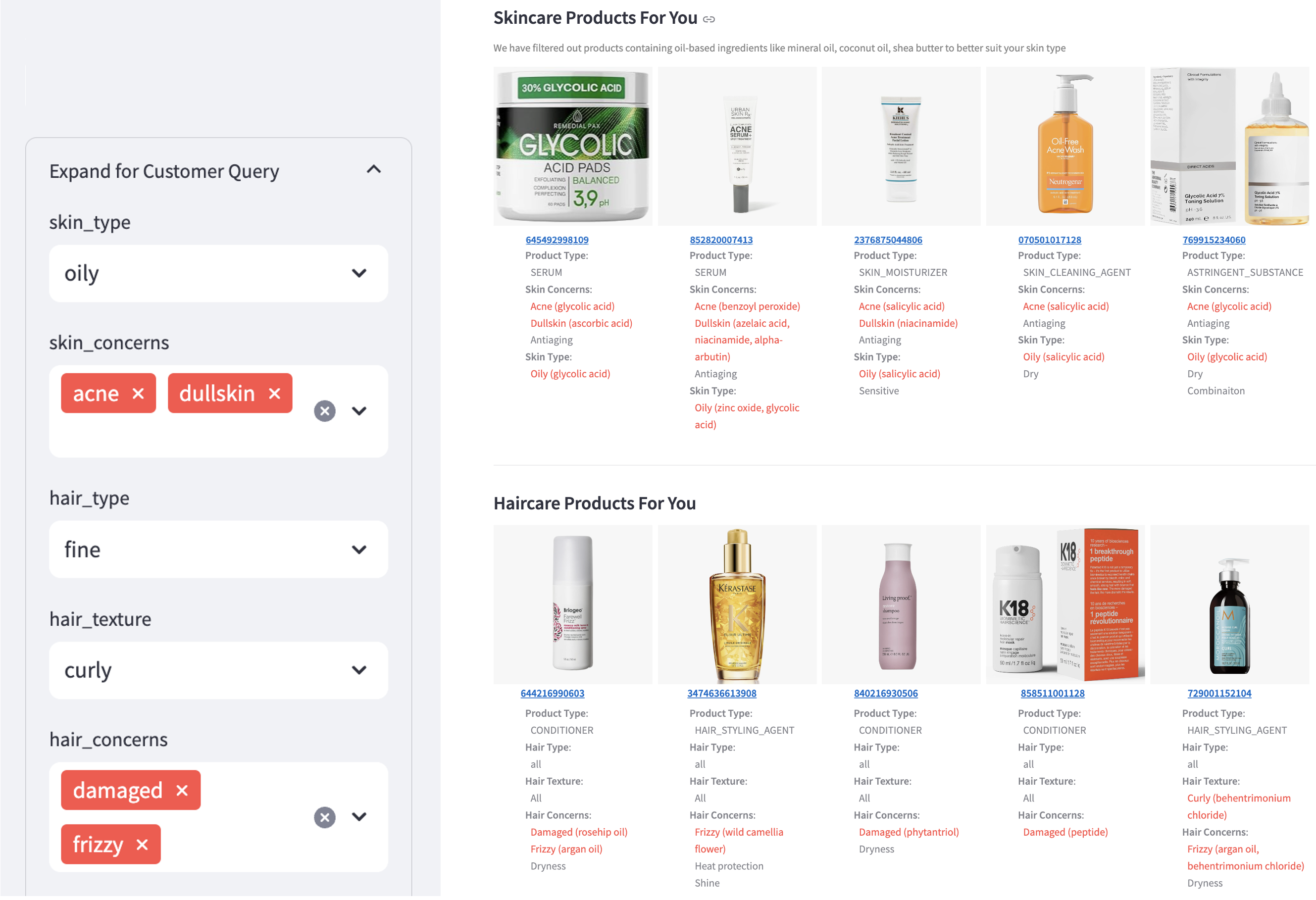}
    \caption{Skincare recommendation with explainable ingredient for each attribute. }
    \label{fig:widget}
\end{figure*}

We approach the beauty attribute extraction problem as a supervised multi-label classification task.
Our proposed solution features a bidirectional Transformer encoder network similar to BERT~\cite{bert}, with a slight modification applied to the last attention layer as summarized in \autoref{algo:forward_pass}.
It is important to note that the network does not use the feed-forward layers in the last Transformer encoder block and does not have any additional classifier modules commonly used in downstream learning tasks. 
Instead, the logits are directly calculated from the attention values.
We refer to our model as \textbf{BeautyTech-BERT}, or \textbf{BT-BERT} for short.

The model operates by taking as inputs a query attribute, a list of ingredients, and the product title, and producing the probability for the query attribute.
\autoref{fig:implicit_diagram} shows an example use-case where the user is querying six attributes for a product titled ``COSRX Snail Mucin Essence". 
Based on the product ingredients, the network will make an inference on whether to label the query attributes true or false. 
In this case, since \texttt{Betaine} is an ingredient known for its hydrating properties, the network is likely to predict true for \texttt{Dry Skin}, meaning this product likely benefits those who have a dry skin type.

Conceptually, our model can be viewed as an energy-based model (EBM)~\cite{ebm2003,51984,lecun2006tutorial}, as it assigns a normalized scalar (or "energy") to each input data point, thereby representing a probability distribution over the training data.
We also denote our model as an \textit{implicit model}, as it accepts the query attribute as input and generates a prediction solely for that attribute.
This distinguishes it from conventional multi-label classifiers, where the classifier module and the number of output classes must be \textit{explicitly} defined.

% Intuitively, our model acts as a skincare expert by learning the correlations between ingredients and the query attributes. It is important to note that the network does not use the feed-forward layers in the last Transformer encoder block and does not have any additional classifier modules commonly used in downstream learning tasks. Instead, the logits are directly calculated from the attention values. Since attention values are bounded between 0 and 1, we scale them by a scaling factor to have a more interpretable range when converting to logits. The scaling factor is a hyper-parameter tuned during training.

% \begin{figure}
%     \centering
%     \includegraphics[width=0.95\textwidth]{Figures/implicit_diagram.pdf}
%     \caption{Overview of beauty product extraction workflow and the BT-BERT architecture. Our model is identical to the BERT Transformer~\cite{bert} except in the last layer---the initial \texttt{N-1} layers remain unmodified. We remove the final MLP from the last layer of the Transformer encoder and directly use the self-attention values to formulate the output probability.}
%     \label{fig:implicit_diagram}
% \end{figure}

\paragraph{Model Input}
For each product, the query attribute is concatenated with ingredients and title to pass to the model. 
Maintaining the original sequence order of the ingredient list is essential, as it reflects the standard convention of listing higher potency ingredients first.
We first tokenize the query label and pad query tokens up to a length of 3. 
The product ingredients and title are also tokenized. 
The entire sequence is truncated or padded such that the final length is 512.
We place the query attribute at the beginning of the input sequence so that its position is consistent across all input sequences---similar to the effect of the {\tt [CLS]} token in BERT when using it in downstream tasks---which is important for computing the logits.

\begin{figure*}
    \centering
    \includegraphics[width=0.8\textwidth]{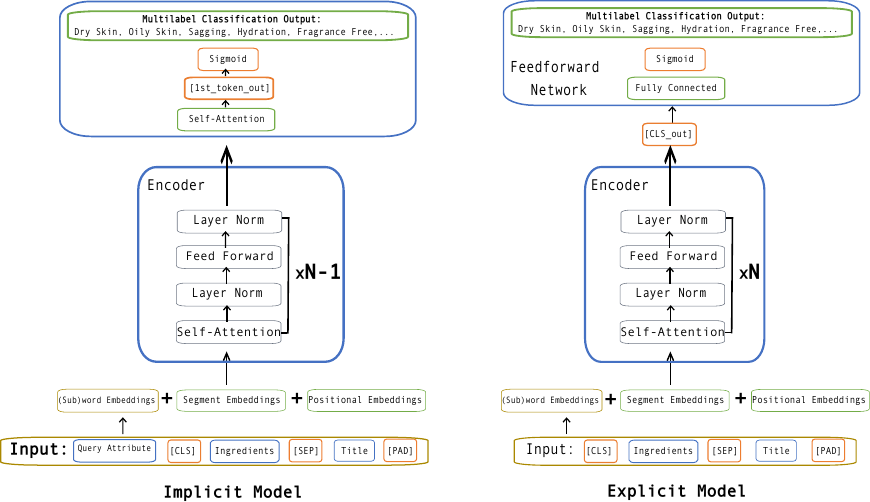}
    \caption{Difference between implicit and explicit models. \textbf{Left:} In implicit models, the model intakes query attribute together with product ingredients and title. Note that in our case, the output logits come directly from the self-attention values of the last encoder layer. 
\textbf{Right:} Explicit models represent the standard way of fine-tuning the BERT model, where a classifier is attached to the end of the Transformer.}
    \label{fig:model_compare}
\end{figure*}

%do we need this?
% Though the code is not complex, I believe this would help others to understand how the last attetion layer is transformed, highlighted in Figure 2.

\begin{algorithm}
   \small
   \caption{\small BT-BERT Forward Pass}
    % \caption{\small SAE Forward Pass}
   \label{algo:forward_pass}
    \begin{algorithmic}[1]
        % \State \texttt{bert\textunderscore model = AutoModel.from\textunderscore pretrained(...) \textcolor{blue}}
        \State \texttt{bert\textunderscore model = AutoModel.from\textunderscore pretrained(...)}
        \State
      \Function{forward}{\texttt{input\textunderscore ids, labels}}
        \State \texttt{outputs = bert\textunderscore model(input\textunderscore ids)}
        \State 
        \State \texttt{\textcolor{blue}{\# extract the last layer's attention, e.g., -1}}
        \State \texttt{\textcolor{blue}{\# attentions are [batch, heads, seqlen, seqlen]}}
        \State \texttt{attentions = outputs["attentions"][-1]}   
        \State 
        \State \texttt{\textcolor{blue}{\# summing attention values over all heads}}
        \State \texttt{\textcolor{blue}{\# for the first token attending to itself}}
        \State \texttt{\textcolor{blue}{\# 16 is a hyperparameter multiplication factor}}
        \State \texttt{logits = 16 * attentions[:, :, 0, 0].sum(dim=1)}
        \State 
        \State \texttt{L = binary\textunderscore cross\textunderscore entropy\textunderscore with\textunderscore logits(logits, labels)}
        \State 
       \EndFunction

\end{algorithmic}
\end{algorithm}
\section{Data Preparation}
\begin{figure}[htbp]
    \centering
    \includegraphics[width=0.48\textwidth]{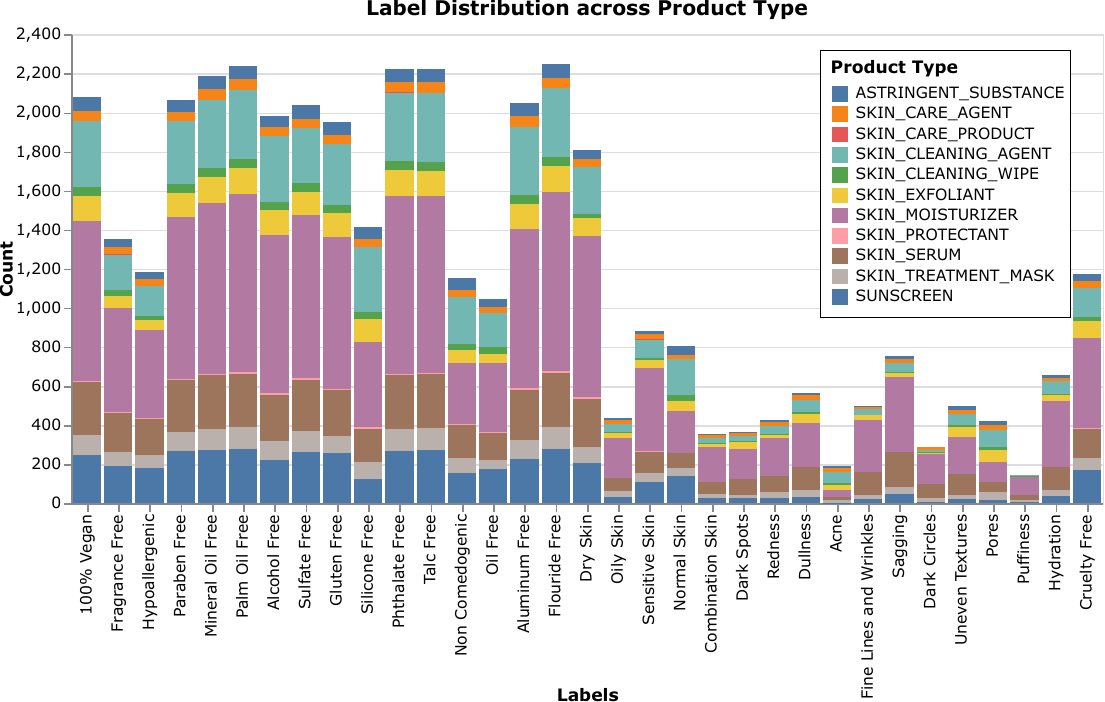}
    \caption{Label Distribution across Product Type in our dataset. The height of each bar indicates the number of products associated with the respective attribute. For instance, there are a total of 1809 out of 11580 products for Dry Skin.}
    \label{fig:label_distribution}
\end{figure}

%should we reallu name the vendor?
Our proposed method is a supervised learning approach and thus requires labeled training data.
% Our dataset is meticulously annotated through rigorous investigation conducted by domain experts in the field of beauty. 
% These annotations are rooted in scientific inquiry and established beauty industry standards, ensuring adherence to regulatory guidelines outlined by the FDA.
% We first gathered a short list of the most frequently searched skincare products. 
We first collect a dataset of skincare products from product data available publicly ~\cite{amazon_ratings_2024,amazon_beauty_2024}.
For each product, attribute labels were meticulously annotated by domain experts based on years of scientific ingredient research.
% In this dataset, each product comes with a list of ingredients and 33 Boolean attribute labels meticulously annotated by domain experts based on years scientific ingredient research. 
An example is shown in \autoref{sec:example_labels}. 
Overall, we collected a total of 11580 data points, where 9334 ($\approx 80\%$) are dedicated to training and 2246 ($\approx 20\%$) to evaluation. 
\autoref{fig:label_distribution} shows the distribution of products categorized by product types and attributes in our dataset. 

% \begin{figure}
%     \centering
%     \includegraphics[width=0.95\textwidth]{Figures/label_distribution.pdf}
%     \caption{Label Distribution across Product Type in our dataset. The height of each bar indicates the number of products associated with the respective attribute. For instance, there are a total of 1809 out of 11580 products for Dry Skin.}
%     \label{fig:label_distribution}
% \end{figure}

\section{Experiments}

This section contains the experiment results and additional analysis around the results.
All experiments are conducted on an EC2 ``p3.16xlarge'' instance with 8 Nvidia Tesla V100 GPUs.

\subsection{Training Details}
For all experiments, we train the network end-to-end with a batch size of 8 until convergence.
We use the AdamW optimizer \cite{adamw-2017} with an initial learning rate of $3\times10^{-5}$.
We follow the standard setup for training Transformer models by splitting the trainable parameters into two categories: decay and non-decay parameters.
Non-decaying parameters are biases and LayerNorm~\cite{ba2016layer} parameters; all other parameters are weight decayed.
We set ${\tt beta2}=0.95$ to improve training stability as recommended in~\cite{zhai2023sigmoid}.

We explored a few different training recipes but found them to have negligible impact on the final model performance, including using a cosine annealing learning rate scheduler~\cite{loshchilov2016sgdr}, linear decay scheduler, and weighted loss for addressing the class imbalance issue. 

\subsection{Baseline Solutions}
We evaluated our method against two simple baseline solutions: Fuzzy Search and the explicit model alternative illustrated in \autoref{fig:model_compare}.

\paragraph{Fuzzy Search}
This is a straightforward approach of finding keywords based on edit distance and other heuristics.
Specifically, a predefined list of target keywords is established (see \autoref{table:fuzzysearch}) for each of the 33 attributes. 
Subsequently, a product is categorized as possessing a particular attribute if any of the keywords from the corresponding list are detected within the product information.

We compare to this baseline as an example of highly explainable solution, but we are well aware that it is not state-of-the-art by any means.
By examining a few examples, the limitations of the fuzzy search approach is immediately apparent.
First, fuzzy search is unable to discern complex textual context.  
For example, it may overlook the labeling of a product described as \textit{free of perfume, silicones, phthalates, fragrance} as `Fragrance Free'.
Second, it is sensitive to error tolerance threshold. 
For instance, despite a product being described as \textit{hydra intensive treatment}, the method may not assign the attribute "Hydration" if the error tolerance is set too low.

\paragraph{Explicit Model}
%how is it easy to add more attributes to the implicit model if it's not trained on them?
A common approach for classification tasks often trains an explicit feed-forward network on top of a pre-trained rich embedding, similar to the approach described in~\cite{bert}. 
As a benchmark, we experimented with this approach, where the model receives product information as input and outputs the likelihood of the 33 labels.  
\autoref{fig:model_compare} highlights the differences between the implicit and the explicit models. 
In the explicit model, the classifier's output dimension is predefined to be the same as the number of attributes.
For this approach, we use the pre-trained weights and tokenizer of {\tt bge-base-en-v1.5}~\cite{bge_embedding} from HuggingFace. 
We chose {\tt bge-base-en-v1.5} as it is considered the state-of-the-art text embedding model for retrieval, clustering, reranking tasks in the Massive Text Embedding Benchmark (MTEB)~\cite{muennighoff2022mteb}.
As a common practice, we freeze the backbone weights and only update the classifier parameters for four epochs to avoid catastrophic forgetting.
We find that training end-to-end after four epochs provides the optimal results compared to other configurations.  

\subsection{Model Results}
%also recall and F1-score?
We evaluate models on the standard classification metrics.
In the following definitions, {\tt TP/TN/FP/FN} refers to the number of true positive, true negative, false positive, and false negative predictions respectively.
\begin{description}
    \item \textbf{Accuracy} is defined as {\tt (TP+TN)/(TP+TN+FP+FN)}.
    \item \textbf{Precision} is defined as {\tt TP/(TP+FP)}. 
    \item \textbf{Recall} is defined as {\tt TP/(TP+FN)}. 
    \item \textbf{F1-Score} is defined as {\tt (2*TP)/(2*TP+FP+FN)}. 
\end{description}

Although we report recall and F1-score, we prioritize accuracy and precision as the main evaluation metrics. 
A higher precision aligns more closely with our acceptable risk threshold by minimizing the likelihood of potentially recommending products containing unsuitable ingredients to customers with particularly sensitive skin.
This is important as we envision attribute-based beauty recommendations as one of the direct applications on this work.
% This choice ensures that our model's predictions of products containing attributes associated with harsh chemicals for treating various skin conditions are highly accurate. 
% Nevertheless, the focus on precision may result in the model overlooking products associated with specific labels.
%, thereby reducing the coverage of the recommender system.

\autoref{tab:explicit-implicit} summarizes the results of label prediction across different methods. 
We observe that both learning-based methods significantly outperform the fuzzy search baseline, as expected.
The implicit model performs slightly better than the explicit alternative across all evaluation metrics.
% Sample products and their predictions are available in \autoref{sec:sample_predictions}.
Aside from the quantitative edge, the implicit model offers other qualitative advantages that the explicit model does not.
We discuss this extensively in the following sections.

%precision is 100?
%Let's also report recall and F1-score
% \begin{table}
% \small
% \caption{Model Performance: Explicit vs.~Implicit Approach (BT-BERT)}
% \renewcommand*{\arraystretch}{1.2}
%   \label{tab:explicit-implicit}
%   \centering
%   \begin{tabular}{lrrrrr}
%     \toprule
%     \textbf{Method} & \textbf{Accuracy} & \textbf{Precision} & \textbf{Recall} & \textbf{F1-Score} & \textbf{Trainable Parameters}\\
%     \midrule
%     BT-BERT & \textbf{0.964} & \textbf{0.987} & \textbf{0.958} & \textbf{0.960} & 109,360,128\\
%     Explicit Model & 0.946 & 0.954 & 0.904 & 0.912 & 109,975,296 \\
%     Fuzzy Search & 0.301 & 0.287 & 0.356 & 0.327 & --\\
%     \bottomrule
%   \end{tabular}
% \end{table}

\begin{table}
\small
\caption{Model Performance: Explicit vs.~Implicit Approach (BT-BERT)}
\renewcommand*{\arraystretch}{1.2}
\label{tab:explicit-implicit}
\centering
\resizebox{0.48\textwidth}{!}{ % Scale table to fit half the column width
  \begin{tabular}{lrrrrr}
    \toprule
    \textbf{Method} & \textbf{Accuracy} & \textbf{Precision} & \textbf{Recall} & \textbf{F1-Score} & \textbf{Parameters}\\
    \midrule
    BT-BERT & \textbf{0.964} & \textbf{0.987} & \textbf{0.958} & \textbf{0.960} & 109,360,128\\
    Explicit Model & 0.946 & 0.954 & 0.904 & 0.912 & 109,975,296 \\
    Fuzzy Search & 0.301 & 0.287 & 0.356 & 0.327 & --\\
    \bottomrule
  \end{tabular}
}
\end{table}

%%%%%%%%%%%%%%%%%%%%%%%%%%%%%%%%%%%%
\subsection{Explainability}
\label{sec:explainability}
% Experiment1: 
In this section, we analyze the input tokens with high attention values in the second last layer of the Transformer encoder block.
Top tokens are obtained using \autoref{algo:attention_values_analysis}.

% We observe that tokens with high attention values are highly relevant to the query attribute.
% Task1: for each label, the tokens show high attention weights make sense for the predicted label. 
In \autoref{table:attention-analysis-by-label}, we choose three query attributes---`Acne', `Fine Lines and Wrinkles' and `Hydration'---and show that tokens with high attention values are ingredients that address the target skin concerns.
This means that our model has learned the effects of different ingredients and how they are associated to different skin concerns and skin types.
We chose these labels, as they are the most popular filter criteria for beauty products. 

% \begin{table}
% \small
% \caption{Attention analysis for `Acne', `Fine Lines and Wrinkles', and `Hydration` attributes}
% \renewcommand*{\arraystretch}{1.2}
%   \label{table:attention-analysis-by-label}
%   \centering
%   \begin{tabular}{lll}
%     \toprule
%     \textbf{Attribute} & \textbf{High Attention Sub-word Tokens} & \textbf{Corresponding Ingredient}\\
%     \midrule
%     \multirow{4}{*}{Acne} & {\tt `sal', `\#ic', `\#yl', `\#ic', `acid'} & Salicylic Acid\\
%     & {\tt `alcohol'} & Alcohol\\
%     & {\tt `benz', `\#oy', `\#l', `per', `\#oxide'} & Benzoyl Peroxide \\
%     & {\tt `beta', `\#ine'} & Betaine \\
%     \midrule
%     \multirow{4}{*}{Fine Lines and Wrinkles} & {\tt `\#pher'} & Tocopheryl Acetate\\
%     & {\tt `\#ito'} & Palmitoyl\\
%     & {\tt `baku', `\#chio'} & Bakuchiol \\
%     & {\tt `re', `\#tino'} & Retinol \\
%     \midrule
%     \multirow{3}{*}{Hydration} & {\tt `\#yal', `\#uron', `ate'} & Sodium Hyaluronate\\
%     & {\tt `\#ly', `\#cer', `\#in'} & Glycerin\\
%     & {\tt `ni', `\#ac', `\#ina', `\#mide'} & Niacinamide \\
%     \bottomrule
%   \end{tabular}
% \end{table}

\begin{table}
\centering
\small
\caption{Attention analysis for `Acne', `Fine Lines and Wrinkles', and `Hydration` attributes}
\renewcommand*{\arraystretch}{1.2}
\resizebox{0.45\textwidth}{!}{ % Resize table to fit half of the column width
  \begin{tabular}{lll}
    \toprule
    \textbf{Attribute} & \textbf{High Attention Sub-word Tokens} & \textbf{Ingredient}\\
    \midrule
    \multirow{4}{*}{Acne} & {\tt `sal', `\#ic', `\#yl', `\#ic', `acid'} & Salicylic Acid\\
    & {\tt `alcohol'} & Alcohol\\
    & {\tt `benz', `\#oy', `\#l', `per', `\#oxide'} & Benzoyl Peroxide \\
    & {\tt `beta', `\#ine'} & Betaine \\
    \midrule
    \multirow{4}{*}{Lines \& Wrinkles} & {\tt `\#pher'} & Tocopheryl Acetate\\
    & {\tt `\#ito'} & Palmitoyl\\
    & {\tt `baku', `\#chio'} & Bakuchiol \\
    & {\tt `re', `\#tino'} & Retinol \\
    \midrule
    \multirow{3}{*}{Hydration} & {\tt `\#yal', `\#uron', `ate'} & Sodium Hyaluronate\\
    & {\tt `\#ly', `\#cer', `\#in'} & Glycerin\\
    & {\tt `ni', `\#ac', `\#ina', `\#mide'} & Niacinamide \\
    \bottomrule
  \end{tabular}
}
\label{table:attention-analysis-by-label}
\end{table}

\begin{table*}
\small
\caption{Attention analysis for product attributes}
\renewcommand*{\arraystretch}{1.2}
\label{table:attention-analysis-by-product}
\centering
\resizebox{0.9\textwidth}{!}{ % Scale table to fit half the column width
  \begin{tabular}{lll}
    \toprule
    \textbf{Attribute} & \textbf{High Attention Sub-word Tokens} & \textbf{Corresponding Ingredient}\\ 
    \midrule
    \midrule
    \multicolumn{3}{p{0.85\textwidth}}{\textbf{Product}: \textit{PanOxyl AM Oil Control Moisturizer, NEW Sheer Formula, Absorbs Excess Oil and Reduces Shine, with Mineral Sunscreen for Acne Prone and Oily And All Skin Tones - 1.7 oz}} \\
    \midrule
    Dry Skin & {\tt `\#yal', `\#uron', `ate'} & Sodium Hyaluronate\\
    Sensitive Skin & {\tt `\#olo'} & Bisabolol\\
    Dark Circles & {\tt `but', `\#yl', `\#ic', `\#yla', `\#te'} & Butyloctyl Salicylate \\
    \midrule
    \multicolumn{3}{p{0.85\textwidth}}{\textbf{Product}: \textit{Good Molecules BHA Clarifying Gel Cream - Facial Cream with Salicylic Acid, Green Tea, and Gotu Kola Extract Soothe and Hydrate - Skincare for Face}} \\
    \midrule
    Acne & {\tt `sal', `\#ic', `\#yl', `\#ic', `acid'} & Salicylic Acid\\
    Dry Skin & {\tt `\#ly', `\#cer', `\#in'} & Glycerin\\
    Redness & {\tt `allan', `\#to'} & Allantoin \\
    \midrule
    \multicolumn{3}{p{0.85\textwidth}}{\textbf{Product}: \textit{I DEW CARE Moisturizer Face Cream - Chill Kitten | Moringa Seed, Prickly Pear, Heartleaf Extract, 24 Hour, Aloe Vera Gel for Dry, Red Skin, Cactus Oil-free, 1.69 Fl Oz}} \\
    \midrule
    Redness & {\tt `tea', `ni', `\#ac', `\#ina', `\#mide'} & Green Tea, Niacinamide\\
    Fine Lines and Wrinkles & {\tt `as', `\#cor', `\#bic'} & Ascorbic Acid\\
    \bottomrule
  \end{tabular}
}
\end{table*}

% task2: per product, show high attention tokens are different for each label prediction
We also assess the high attention tokens for each predicted label of a single product and show that these tokens are different across attributes of a given product.
This means that our model has learned to pay attention to different tokens when it is being asked about different attributes.
\autoref{table:attention-analysis-by-product} demonstrates some of the examined products.

\subsection{Robustness in Low Data Regime}
\label{sec:robustness}

% \subsection{Learning Curves for Robustness in Low Data Regime Experiment}

\begin{figure}
    \centering
    \includegraphics[width=0.45\textwidth]{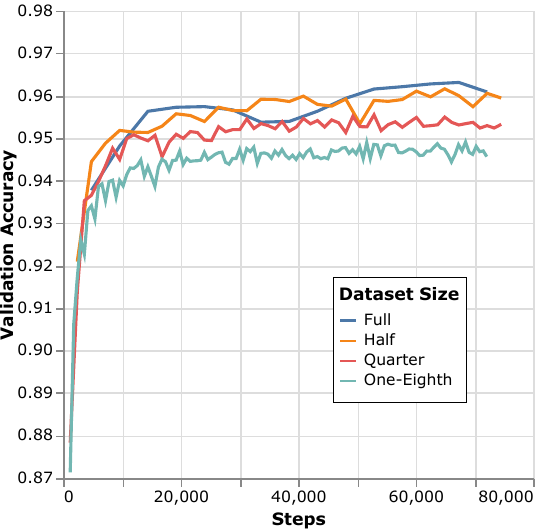}
    \caption{Validation accuracy training on various sizes of dataset}
    \label{fig:low_data}
\end{figure}

% In this section, we show BT-BERT performs well even under low-resource scenarios. 
In this section, we present empirical evidence demonstrating the robust performance of BT-BERT even when the volume of training data is limited.
\autoref{fig:low_data} shows the validation accuracy across various degrees of data scarcity, namely when the model is trained using the full dataset, as well as 1/2, 1/4, and 1/8 of the full training corpus. 
In each training run, we systematically down-sample the training set and keep the validation set constant, i.e., it still contains the same 2246 products.

Note that for the 1/8 training, the model is trained with only 1167 products and yet still the validation accuracy only drops by less than 1.25\%. 
We hypothesize that the robust performance of BT-BERT in such a low-resource regime can be attributed to the fact that it is an energy-based implicit model, as opposed to an explicit classifier.
The same scaling pattern is observed in other energy-based models \cite{implicit-2021}.
Additionally, we attribute part of such robustness to the implicit data augmentation strategy employed in training---specifically, each product is paired with all 33 query attributes, exposing our model to diverse input contexts.
We have not yet fully characterize the scaling behaviors of implicit and explicit models. 
It is possible that with improved training techniques, the explicit approach can close the gap in low-resource regimes.
\section{Discussion}

\subsection{Does the choice of logits transformation matter?}
Our early experiments indicate that scaling the probability linearly with 16 achieves better results than not employing it. 
% experiment v137 (log-12 head 1 token) vs v91(12 head 1 token)
We explored an alternative scaling formulation using ${f(x) = \log(x / (1-x))}$, where $x$ represents the attention value of the first query token from all attention heads.
The design is inspired by probability theory, where $x / (1-x)$ is commonly referred to as the odds or odds ratio when $x$ is a probability.
Taking the logarithm of the odds ratio is a common transformation used in logistic regression to convert probability into logits.

% The design of this log transformation is to preserve the information of the original signal.
% experiment v90 (log-12 head 3 token sum), v85(log-12 head 3 token mean)
Additionally, we experimented with using the summation and average of the attention values from the first three query tokens as $x$ before applying the log transformation. 
However, these variations did not produce better results. 
Ultimately, we chose the linear scaling method of multiplying by 16 due to its simplicity and slightly faster computation times.
% We hypothesize the range of probabilities in our task can be modeled by a fixed range of values and can benefit from the efficiency of linear transformations.
% This is talking about using heads.. not so related to this section
% 12 head 1 token > 12 head 3 sum > 1 head 1 token > 12 head 3 mean
% We observed that using attention values from all heads is generally better than that of 1 head.
% We hypothesize that this due to the diverse insights harnessed by each attention head in the transformer architecture.

\subsection{Finetuning on Additional Attributes}
\label{sec:fine-tune}
% In this experiment, we test the capability of an implicit model for its ability to adapt when new labeled attributes become available.
% We consider a scenario where the dataset was initially released with only 30 out of the 33 labels, and the remaining 3 labels were made available in the second release.
In this section, we discuss the adaptability of implicit models in incorporating new labeled attributes as they become available. 
We design a scenario mirroring real-world dynamics, where an initial dataset comprises 30 out of 33 labels, with the remaining 3 labels introduced in a subsequent release. 
Such scenarios are commonplace in the beauty industry, where emerging trends and evolving consumer preferences necessitate the addition of new product attributes. 
For instance, the advent of clean beauty as a trend in 2023~\cite{clean_beauty} underscores the relevance of this work.
Through comprehensive analysis and experimentation, we assess and highlight the implicit model's efficacy in seamlessly incorporating new attributes.

% Without a loss of generality, we run this experiment with one-eighth of the full training dataset, since \autoref{fig:low_data} indicates minimal performance degradation compared to training on the full dataset.
We removed the labels for `Fragrance Free' (generally-preferred), `Oily Skin' (skin type), and `Acne' (skin concern) from the full dataset ($\mathcal{D}_{\text{full}}$) and trained a model on the remaining 30 labels ($\mathcal{D}_{\text{30}}$).
Then, we add back the removed labels and finetune the previously trained model with the complete dataset for only one epoch.
\autoref{table:flexibility-results} shows the validation accuracies before and after the finetuning step.
When finetuning on only the three additional labels ($\mathcal{D}_{\text{3}}$), we observe a significant drop in validation accuracy for the existing 30 labels in the validation set.
We believe this is due to the \textit{catastrophic forgetting} problem and could potentially be alleviated by using more advanced finetuning algorithms~\cite{hu2021lora,liu2024dora,zhang2023adding}.

When finetuning with $\mathcal{D}_{\text{full}}$, we observe only a slight drop of performance when predicting the existing 30 labels, but the accuracy for the new labels is drastically improved.
It is important to note that this finetuning procedure is impossible when using explicit models, since the number of output classes is different and therefore the classifier must be replaced and retrained.
% \begin{table}
%   \centering
%   \small
%   \caption{Model performance on partially held out data. In this experiment, we evaluate the model's ability to incorporate additional labels when they become available.}
%   \label{table:flexibility-results}
%   \begin{tabular}{lc|cc}
%     \toprule
%     & Train with $\mathcal{D_\text{30}}$ & Finetune with $\mathcal{D_\text{3}}$ & Finetune with $\mathcal{D_\text{full}}$ \\
%     \midrule
%     Acc.~on 30 labels & 93.9\% & 82.4\% & 93.4\% \\
%     Acc.~on 3 labels & 59.6\% & 94.7\% & 93.5\% \\
%     \bottomrule
%   \end{tabular}
% \end{table}

\begin{table}
  \centering
  \small
  \caption{Model performance on partially held out data. In this experiment, we evaluate the model's ability to incorporate additional labels when they become available.}
  \label{table:flexibility-results}
  \resizebox{0.48\textwidth}{!}{ % Scale table to fit half the column width
    \begin{tabular}{lc|cc}
      \toprule
      & Train $\mathcal{D_\text{30}}$ & Finetune $\mathcal{D_\text{3}}$ & Finetune $\mathcal{D_\text{full}}$ \\
      \midrule
      Acc.~on 30 labels & 93.9\% & 82.4\% & 93.4\% \\
      Acc.~on 3 labels & 59.6\% & 94.7\% & 93.5\% \\
      \bottomrule
    \end{tabular}
  }
\end{table}

\begin{algorithm}
   \footnotesize
   \caption{\small Key Token Extraction Based on Attention Values}
   \label{algo:attention_values_analysis}
    \begin{algorithmic}[1]
      \Function{GetTopAttentionTokens}{\texttt{input\textunderscore ids, attentions, topk}}
        \State \texttt{\textcolor{blue}{\# input\textunderscore ids is a tensor of shape (seqlen,)}}
        \State \texttt{\textcolor{blue}{\# attentions is a tensor of shape (heads, seqlen, seqlen)}}
        \State
        \State \texttt{\textcolor{blue}{\# get index of top-k attention per row across all heads}}
        \State \texttt{topk\textunderscore indices = attentions.flatten(0, 1).topk(topk).indices}
        \State \texttt{topk\textunderscore indices = topk\textunderscore indices.unique()}
        \State 
        \State \texttt{\textcolor{blue}{\# convert col indices to token strings}}
        \State \texttt{topk\textunderscore tokens = convert\textunderscore ids\textunderscore to\textunderscore tokens(input\textunderscore ids[topk\textunderscore indices])}
        \State
        \State \texttt{\textcolor{blue}{\# remove non-meaningful tokens}}
        \State \texttt{TO\textunderscore REMOVE = [`,', `[CLS]', `[SEP]', `(', `)', `[PAD]']}
        \State \texttt{topk\textunderscore tokens = [k for k in topk\textunderscore tokens if k not in TO\textunderscore REMOVE]}
        \State  
       \EndFunction

\end{algorithmic}
\end{algorithm}

\subsection{Alternating Query Attribute Tokens}
% flexibility of which query label token, what are the possible prod senario?
% would this belong to explainability or other secitons?
% task3: When changing input label token, how does the high attention token changes
In this section, we highlight the benefit of our implicit model during inference time.
First, we show that it can handle similar but not identical query attributes.
% We assess the flexibility of our implicit model in adapting to similar query attribute alternatives, and show that the tokens with high attention values using\autoref{algo:attention_values_analysis} generated from two similar query attributes are almost identical.
% We take `Fine Lines and Wrinkles' as an example and replace the query attribute with just a single word `Lines' for product \textit{DERMA-E Anti-Wrinkle Renewal Skin Cream}.
We take `Fine Lines and Wrinkles' as an example and replace the query attribute with just a single word `Lines' for a commonly available anti-wrinkle renewal skin cream.
We use \autoref{algo:attention_values_analysis} to extract the high attention tokens and track how they change when the attribute tokens are replaced.

We observed a number of overlapping tokens especially those addressing lines and wrinkles---{\tt `\#chio'}, {\tt `pu'}, {\tt `soy'}, {\tt `lines'}, {\tt `baku'}, {\tt `re'}, and {\tt `\#tino'}.
We also identified non-overlapping tokens such as {\tt water}, {\tt after}, {\tt cleansing}, {\tt fine}, {\tt cart}, and {\tt wr}.
% It is important to note that the non-overlapping tokens are more general and not as directly relevant in addressing the particular skin concern, e.g., water and cleansing. 
It is important to note that the non-overlapping tokens, such as 'water' and 'cleansing,' are more general and not as directly relevant to the specific skin concern.
We believe that this approach can help us better understand the ingredients and their target uses. 

% Specifically, using `Fine Lines and Wrinkles', tokens with high attention values are {\tt 'fine', 'lines', 'wr', 'pu', '\#ius', 'sa', 'ste', '\#ari', 'micro', '\#cr', 'baku', '\#chio', '\#l', 'soy'}; while using `Lines', tokens with high attention values are {\tt 'lines', 'pu', 'water', 'cart', '\#ham', '\#us', '\#ius', 'sa', 'micro', '\#cr', 'baku', '\#chio', '\#l', 're', '\#tino', '\#l', 'after', 'cleansing', 'soy'}
% - Using `Fine Lines and Wrinkles' :['fine', 'lines', 'wr', 'pu', '\#ius', 'sa', 'ste', '\#ari', 'micro', '\#cr', 'baku', '\#chio', '\#l', 'soy']
% - Using `Lines': ['lines', 'pu', 'water', 'cart', '\#ham', '\#us', '\#ius', 'sa', 'micro', '\#cr', 'baku', '\#chio', '\#l', 're', '\#tino', '\#l', 'after', 'cleansing', 'soy']
% There are 10 tokens overlap, which is almost identical.

% \subsection{\note{flexibility of which label to predict}}
% does this subsection make this section inconsistent?
% Experiment5: just a beautiful algorithm, not much useful in prod

\section{Explainable Beauty Recommendation and Customer Understanding}
\label{sec:customer-segmentation}

% \begin{figure*}
%     \centering
%     \includegraphics[width=0.9\textwidth]{Figures/widget.png}
%     \caption{Skincare recommendation with explainable ingredient for each attribute. }
%     \label{fig:widget}
% \end{figure*}

\paragraph{\bf Explainable Beauty Recommendation}  One critical application of ingredient-based attribute extraction lies in delivering explainable recommendations to beauty customers. 
%% Why explainable is important 
In the ever-evolving beauty industry, where personalization is key, transparency and clarity in product suggestions are vital.
As illustrated in \autoref{fig:widget}, skincare recommendations are made using a point-wise approach, where each product is individually assessed based on the customer's specific skin type and concerns.
Here, the customer has selected ``oily'' skin and concerns of ``acne'' and ``dullskin''. 
% The recommended products not only address these issues but are also compatible with the customer’s skin type, enhancing the trustworthiness and relevance of each suggestion.
% Each product is annotated with its predicted target skin concerns and skin types, alongside the active ingredients responsible for these benefits, using \autoref{algo:attention_values_analysis} discussed in \autoref{sec:explainability}.
% For example, Salicylic Acid is highlighted for its anti-acne properties across various product types like cleansers, pads, and serums. 
% based on tracy's feedback
The recommended products not only contain ingredients intended to address these issues but are also compatible with the customer’s stated skin type, enhancing the trustworthiness and relevance of each suggestion. 
Each product is annotated with its predicted target skin concerns and skin types, alongside the ingredients intended to address those concerns, using \autoref{algo:attention_values_analysis} discussed in \autoref{sec:explainability}.
For example, Salicylic Acid is highlighted for its anti-acne properties across various product types like cleansers, pads, and serums. 
% Furthermore, the system strategically omits products with oil-based ingredients that could exacerbate oily skin, ensuring that recommendations are both effective and appropriate for the user's concerns.
Furthermore, the system strategically omits products with oil-based ingredients that could exacerbate oily skin, ensuring that recommendations are appropriate for the user’s concerns.

% By providing fact-based explanations for the benefits of recommended products, this approach offers clear and transparent justifications for the recommendations.
By providing fact-based explanations for recommended products, this approach offers clear and transparent justifications for the recommendations.
As customers purchase and use products with effective ingredients, they are more likely to achieve the desired skin results, fostering long-term trust and encouraging repeat engagement with the e-commerce store.
This method not only empowers customers to make informed purchasing decisions but also strengthens their trust in the recommendation system. 
This approach is versatile and can be applied broadly across most beauty catalogs, including haircare and makeup, where ingredients stay on the skin for extended periods.
In the context of strategic and utility-aware recommendations, explainability is crucial for aligning personalized suggestions with both individual needs and broader objectives. 
This alignment ultimately enhances customer confidence, satisfaction, and long-term audience growth.
% Explainability plays a vital role in aligning personalized recommendations with both individual preferences and broader goals, ultimately fostering customer confidence, satisfaction, and sustained engagement.

% \begin{figure*}
%   \centering
%   \includegraphics[width=.9\linewidth]{Figures/segments.pdf}
%   \caption {Cohorts of customers having low and high propensities towards `Acne' and `Fine Lines and Wrinkles', and their overlap. }
%   \label{fig:segment}
% \end{figure*}

% One of the critical use cases of product attribute extraction is building the implicit beauty customer profile. An implicit customer profile refers to a list of preferences and concerns each customer has towards beauty products, reflected in their past purchases. Having this customer profile enables us to personalize recommendations and create segments of customers.
\paragraph{\bf Customer understanding} Conversely, customer propensity toward specific attributes—such as preferred skin type, skin concerns, and ingredient preferences—can be inferred from their past purchases.
Our future work focuses on understanding customer skin types and concerns by building upon existing attribute extraction methodologies. 
This advancement will enable further refinement of our recommendation algorithms, particularly in the ranking layer.

\section{Conclusion}

% The explainability, flexibility, and scability benefits of implicit models apply in general to all machine learning applications.
% However, in the work, we focus our experiments and analysis on the beauty domain.

We present an energy-based implicit model for extracting beauty-specific attributes trained using end-to-end supervised learning.
We empirically show that the implicit approach outperforms traditional explicit classifiers in terms of accuracy, precision, and other evaluation metrics.
Aside from better performance, we show that the implicit model is explainable, robust to low-data scenarios, and easy to incorporate new attributes as they become available.
Using the explainability feature of our model, we propose novel ways to use the predictions without additional training by comparing and contrasting the high value tokens across different products and attributes.
We have not yet fully characterized the limits of the model’s capabilities.
Currently, we only qualitatively identify the high attention value tokens and discuss how they are related to the specific skin concerns and skin types in our attention analysis. 
We wish to better quantify the correlations between all predicted ingredients and the attributes.
Although our work focuses on beauty attribute extraction, we believe the simplicity of our approach and comprehensiveness of our analysis provide a solid foundation for future research in designing more capable and explainable models in all domains of machine learning.
In future work, we will validate the generated attributes within downstream recommendation systems and conduct a thorough evaluation. Furthermore, we will assess the impact of explainability for end users through A/B testing.

%%
%%%%%%%%% REFERENCES
\bibliographystyle{ACM-Reference-Format}
\bibliography{references}

\appendix
% {\LARGE Supplementary Materials}
\section{Appendix}
\subsection{Labels for Skincare Products}
\label{sec:label_category}
We define 33 labels for skincare products that include 5 skin types, 11 skin
concerns, and 17 attributes that are generally preferred across beauty products.
\begin{itemize}
    \item Target skin types: Dry Skin, Normal Skin, Oily Skin, Combination Skin, Sensitive Skin
    \item Target skin concerns: Acne, Hydration, Pores, Fine Lines and Wrinkles, Sagging, Dark Spots, Dullness, Redness, Uneven Texture, Dark Circles, Puffiness
    \item General preferred beauty attributes: 100\% Vegan, Cruelty Free, Fragrance Free, Hypoallergenic, Paraben Free, Mineral Oil Free, Palm Oil Free, Oil Free, Alcohol Free, Sulphate Free, Gluten Free, Silicone Free, Phthalate Free, Talc free, Non Comedogenic, Aluminum Free, Fluoride Free.
\end{itemize}

\subsection{Product information and Labels}
\begin{figure*}[htbp]
    \centering
    \includegraphics[width=0.9\textwidth]{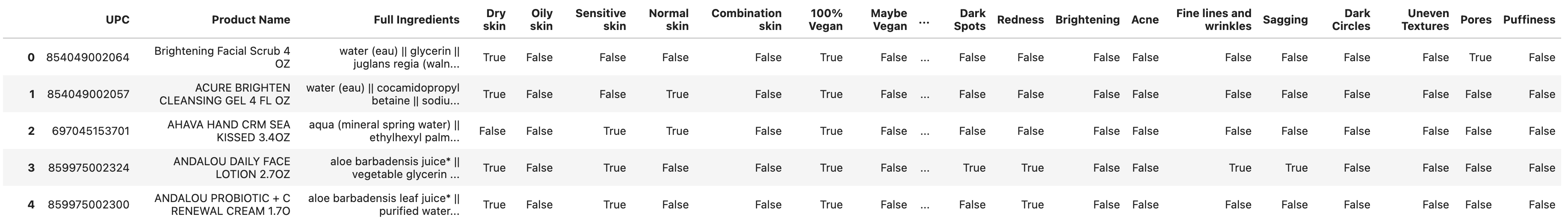}
    \caption{Sample Pandas dataframe with product ingredient list (\texttt{Full Ingredients}) and title (\texttt{item\_name)} for each product.}
    \label{sec:example_labels}
\end{figure*}

Each product comes with a title, list of ingredients, and a Boolean label for each attribute. An example is shown in \autoref{sec:example_labels}.

% \begin{figure}[h]
%     \centering
%     \includegraphics[width=0.9\textwidth]{Figures/example_labels.pdf}
%     \caption{Sample Pandas dataframe with product ingredient list (\texttt{Full Ingredients}) and title (\texttt{item\_name)} for each product.}
%     \label{sec:example_labels}
% \end{figure}

\subsection{FuzzySearch Attribute Key Words}
\label{table:fuzzysearch}
For FuzzySearch method, We define keywords for each of the 33 labels.
\begin{itemize}
    \item Dry Skin: "dry", "all", "universal".
    \item Normal Skin: "normal", "all", "universal".
    \item Oily Skin: "oil", "all", "universal".
    \item Combination Skin: "combination", "all", "universal".
    \item Sensitive Skin: "sensitive", "all", "universal".
    \item Acne: 
    "anti acne",
    "blackheads",
    "salicylic acid",
    "Glycolic Acid",
    "Benzoyl Peroxide",
    "breakouts treatment",
    "acne preventing",
    "skin clarifying".
    \item Hydration: "dehydration",
    "dryness",
    "hydrating",
    "rehydrate",
    "soothing",
    "moisturizing",
    "nourishing",
    "softening",
    "replenishing".
    \item Pores: "pore", "oil control".
    \item Fine Lines and Wrinkles: "wrinkle",
    "anti-aging",
    "anti aging",
    "anti-aging",
    "wrinkle treatment",
    "wrinkles treatment",
    "skin cell renewal",
    "skin-cell-renewal",
    "plumping",
    "refine skin texture",
    "refine-skin-texture",
    "repairing",
    "fine line",
    "anti aging",
    "plumping",
    "skin cell renewal",
    "replenishing",
    "octinoxate",
    "octisalate",
    "avobenzone".
    \item Sagging: "firming", "wrinkle", "anti aging", "skin cell renewal".
    \item Dark Spots: "hyperpigmentation",
    "melasma",
    "dyschromia",
    "brown spot",
    "age spot",
    "dark spot",
    "brightening",
    "even toning",
    "color correction",
    "lightening",
    "antioxidant",
    "oxygenating",
    "whitening".
    \item Dullness: "even toning",
    "dull skin",
    "lightening",
    "brightening",
    "colour correction",
    "skin cell renewal",
    "rejuvenating",
    "exfoliating",
    "plumping".
    \item Redness: "redness", "anti inflammatory", "soothening", "soothing", "redness reduction", "redness removal", "oxygenating".
    \item Uneven Texture: "uneven texture", "uneven skin".
    \item Dark Circles:"puffiness",
    "dark circles",
    "color correction",
    "lightening",
    "antioxidant",
    "radiant skin",
    "brightening".
    \item 100\% Vegan: "vegetarian", "plantbased", "vegan", "animalbyproductfree".
    \item Cruelty Free: "crueltyfree".
    \item Fragrance Free: "unscented", "fragrancefree". 
    \item Hypoallergenic: "preservativefree",
    "latexfree",
    "chemicalfree",
    "formaldehydefree",
    "slesfree".
    \item Paraben Free: "preservativefree", "slesfree", "slsfree", "parabenfree".
    \item Mineral Oil Free: "palmoilfree", "mineraloilfree".
    \item Palm Oil Free: "palmoilfree".
    \item Oil Free: "oilfree", "palmoilfree", "mineraloilfree".
    \item Alcohol Free: "alcoholfree". 
    \item Sulphate Free: "sulfatefree".
    \item Gluten Free: "glutenfree". 
    \item Silicone Free: "siliconefree". 
    \item Phthalate Free: "phthalatefree".
% https://code.amazon.com/packages/BeautyTechRecommender/blobs/bparida-dev/--/src/beauty_tech_recommender/training/product_profile.py
\end{itemize}

\end{document}